\documentclass[a4paper, 10pt, conference]{ieeeconf}      % Use this line for a4
\usepackage{FG2026}
\usepackage{amsmath}
\usepackage{amsmath,amssymb,amsfonts}%
\usepackage{mathrsfs}%
\usepackage{mathtools}
\usepackage{todonotes}
\usepackage{float}
\usepackage{algorithm}
\usepackage{algpseudocode}
\usepackage{booktabs}
\usepackage{tabularx}

\usepackage{cleveref}

\renewcommand{\paragraph}[1]{\vspace{1em}\noindent\textbf{#1}.}

\AtBeginDocument{%
}

\def\etal{\textit{et al.}}

\FGfinalcopy % *** Uncomment this line for the final submission

\IEEEoverridecommandlockouts                              % This command is only
\def\FGPaperID{264} % *** Enter the FG2026 Paper ID here

\title{\LARGE \bf
A Non-Invasive 3D Gait Analysis Framework for Quantifying Psychomotor Retardation in Major Depressive Disorder
}

\author{Fouad Boutaleb$^{1,4}$, Emery Pierson$^{2}$, Mohamed Daoudi$^{1,3}$, Clémence Nineuil$^{4}$, Ali Amad$^{4}$, Fabien D'Hondt$^{4}$\\
    $^1$ Univ. Lille, CNRS, Centrale Lille, Institut Mines-Télécom, UMR 9189 CRIStAL, F-59000 Lille, France\\
    $^2$ LIX, École Polytechnique, IP Paris\\
    $^3$ IMT Nord Europe, Institut Mines-Télécom, Univ. Lille, Centre for Digital Systems, F-59000 Lille, France\\
    $^4$Univ. Lille, Inserm, CHU Lille, U1172 - LilNCog - Lille Neuroscience \& Cognition, F-59000 Lille, France\\
    }
\begin{document}

\ifFGfinal
\thispagestyle{empty}
\pagestyle{empty}
\else
\author{Anonymous FG2026 submission\\ Paper ID \FGPaperID \\}
\pagestyle{plain}
\fi
\maketitle

%%%%%%%%%%%%%%%%%%%%%%%%%%%%%%%%%%%%%%%%%%%%%%%%%%%%%%%%%%%%%%%%%%%%%%%%%%%%%%%%
\begin{abstract}

Predicting the status of Major Depressive Disorder (MDD) from objective, non-invasive methods is an active research field. Yet, extracting automatically objective, interpretable features for a detailed analysis of the patient state remains largely unexplored.
Among MDD's symptoms, Psychomotor retardation (PMR) is a core item, yet its clinical assessment remains largely subjective. While 3D motion capture offers an objective alternative, its reliance on specialized hardware often precludes routine clinical use. In this paper, we propose a non-invasive computational framework that transforms monocular RGB video into clinically relevant 3D gait kinematics. Our pipeline uses Gravity-View Coordinates along with a novel trajectory-correction algorithm that leverages the closed-loop topology of our adapted Timed Up and Go (TUG) protocol to mitigate monocular depth errors. This novel pipeline enables the extraction of 297 explicit gait biomechanical biomarkers from a single camera capture. 

To address the challenges of small clinical datasets, we introduce a stability-based machine learning framework that identifies robust motor signatures while preventing overfitting. Validated on the CALYPSO dataset, our method achieves an 83.3\% accuracy in detecting PMR and explains 64\% of the variance in overall depression severity ($R^2=0.64$). Notably, our study reveals a strong link between reduced ankle propulsion and restricted pelvic mobility to the depressive motor phenotype. These results demonstrate that physical movement serves as a robust proxy for the cognitive state, offering a transparent and scalable tool for the objective monitoring of depression in standard clinical environments. 

\end{abstract}

%%%%%%%%%%%%%%%%%%%%%%%%%%%%%%%%%%%%%%%%%%%%%%%%%%%%%%%%%%%%%%%%%%%%%%%%%%%%%%%%
\section{INTRODUCTION}
MDD is a common and disabling psychiatric condition, with a lifetime prevalence of approximately 10\% worldwide~\cite{marx2023major}. Beyond its high prevalence, MDD is associated with substantial functional disability and increased mortality, including suicide risk. Importantly, MDD is marked by pronounced clinical heterogeneity,  encompassing diverse symptom profiles, illness trajectories, and treatment responses. This heterogeneity has raised major concerns regarding the limits of current diagnostic categories and has driven efforts to identify more informative clinical dimensions and phenotypes to better characterize, stratify, and understand depressive disorders~\cite{allsopp2019heterogeneity, amad2025peripheral},~\cite{boualeb2025measuring}.

Among these dimensions, PMR stands out as a core clinical feature of MDD, encompassing both cognitive and motor slowing. Its motor manifestations—such as reduced movement initiation, altered gait, postural changes, and diminished expressive behavior—are particularly amenable to objective quantification~\cite{bennabi2013psychomotor}. PMR is frequently observed in moderate to severe depression and is strongly associated with illness severity and functional impairment~\cite{bennabi2013psychomotor, douka2025cognitive}. Moreover, it carries prognostic value, as it has been shown to predict treatment response and recovery trajectories~\cite{terziivanova2018objective, paquet2022psychomotor}.

In clinical practice, however, PMR is rarely quantified using dedicated instruments and is most often assessed through a global clinical impression, based on the clinician's subjective perception of motor slowing during the interview. While such impressions are central to psychiatric evaluation, they are inherently limited by rater variability.
%, contextual influences, and reduced sensitivity to subtle or progressive changes over time, limiting the use of PMR as a reliable and sensitive marker of illness severity, treatment response, and longitudinal course. 
Although specific clinical tools targeting psychomotor disturbance exist, their use remains limited outside research settings due to time constraints and feasibility issues in routine care~\cite{paquet2022psychomotor, song2025assessing}.

In this context, gait represents a particularly relevant motor behavior to investigate. As a global and ecologically valid motor activity, gait captures core dimensions of psychomotor functioning, including initiation, speed, posture, and overall motor organization, which are frequently altered in depressive psychomotor slowing~\cite{ActaPsychologica2025}. Unlike isolated or purely volitional motor tasks, gait reflects an integrated pattern of motor behavior that closely aligns with the clinician’s global impression of PMR. This makes gait a suitable candidate for the objective characterization of PMR, allowing the extraction of quantitative motor features that are less susceptible to subjective bias and effort-related variability than traditional observational assessments.

This study proposes a reliable, non-invasive computational framework for quantifying PMR, specifically focusing on the automated extraction of gait kinematics from monocular video. By shifting from qualitative clinical observation to objective analysis, we provide a scalable methodology for capturing the biomechanical signatures of psychomotor impairment as digital biomarkers. Specifically, we adapt the TUG protocol~\cite{podsiadlo1991tug} to our needs, and integrate a state-of-the-art monocular 3D human reconstruction method (GVHMR)~\cite{shen2024gvhmr}, with a trajectory-correction algorithm to correct drift from monocular human body estimation. We extract clinically relevant features that we validate in a cohort of 42 patients from the CALYPSO dataset. Our approach offers an objective and scalable solution that provides objective decision support in clinical settings. 

\noindent Our main contributions are summarized as follows:
\begin{itemize}
    \item A standardized acquisition protocol based on a modified TUG test, enabling reproducible gait capture and objective depression assessment in standard clinical settings.
    \item A novel monocular 3D human motion capture framework. Based on Gravity-View Coordinates (GVHMR)~\cite{shen2024gvhmr}, we use a novel trajectory-correction algorithm that leverages the closed-loop topology of our protocol to mitigate monocular drift — improving severity estimation by over 40\%.
    \item An automated pipeline extracting 297 biomechanical parameters (spatiotemporal, kinematic, arm swing) that directly map to psychiatric semiology, such as stride length and pelvic mobility.
    \item A stability-based machine learning framework that robustly identifies motor signatures despite small sample sizes, achieving 83.3\% accuracy in detecting psychomotor retardation and an $R^2$ of 0.64 for depression severity.
\end{itemize}

%%%%%%%%%%%%%%%%%%%%%%%%%%%%%%%%%%%%%%%%%%%%%%%%%%%%%%%%%%%%%%%%%%%%%%%%%%%%%%%%

\section{RELATED WORK}

\subsection{Clinical Assessment of Psychomotor Retardation}
PMR is primarily evaluated during the clinical interview and is rarely quantified in routine practice. In most cases, its assessment relies on the clinician’s global impression of motor slowing, integrating observations of movement, posture, expressiveness, and overall psychomotor behavior, without the use of standardized metrics~\cite{bennabi2013psychomotor, paquet2022psychomotor}.

Standardized clinical assessments exist; some rely on simple items embedded within broader depression rating scales (e.g., Hamilton Depression Rating Scale (HDRS) ~\cite{hamilton1960hdrs}), while others are specifically designed to assess psychomotor retardation, such as the Salpêtrière Retardation Rating Scale~\cite{widlocher1983psychomotor} or the CORE index~\cite{parker1994defining}. However, their use remains largely confined to research settings due to time constraints and limited feasibility in everyday clinical care~\cite{song2025assessing}. These limitations motivate objective measurement approaches aimed at providing standardized and sensitive assessments of psychomotor functioning.

\subsection{Instrumental and Sensor-Based Approaches}
The gold standard for quantifying gait kinematics remains marker-based optical motion capture (MoCap) systems~\cite{whittle2014gait}. However, their high cost and spatial requirements render them inaccessible for routine psychiatry.

To address these barriers, researchers adopted actigraphy and IMU-derived metrics~\cite{burton2013actigraphy,maruani2025actigraphy,ouzar2025wearable}. While effective for measuring global activity levels, these summaries lack the granularity to characterize specific joint dynamics. Moreover, wearable hardware is intrusive for sensitive patients and imposes maintenance burdens that limit scalability.

Alternatively, depth sensors (e.g., Microsoft Kinect) were explored to capture skeletal data without body instrumentation~\cite{SciRepGait2025,daoudiaciiw2019}. While these systems offer a non-invasive solution for measuring gross motor activity, they suffer from critical hardware-imposed limitations: their restricted capture volume precludes the analysis of extended protocols like the TUG, and they rely on specific, often discontinued sensors. This hardware dependency creates a significant deployment barrier compared to standard video-based solutions.

\subsection{Video-based Assessment}
To overcome the hardware dependencies of depth sensors and wearables, recent research has pivoted toward purely RGB-based computer vision methods.

Multi-view systems such as OpenCap~\cite{uhlrich2023opencap, Min_2024} attempt to replicate MoCap precision using synchronized smartphones. While validated for neurological disorders~\cite{Xu_2025}, the requirement for multi-device calibration limits broad clinical adoption.

This establishes monocular 3D human mesh recovery exemplified by models like Video Inference for Body
Pose and Shape Estimation (VIBE) \cite{kocabas2020vibe} and VisionMD-Gait \cite{Liu2026VisionMD}—as the most scalable alternative. However, a critical gap remains: these models are currently restricted to disorders with ``visible'' or coarse motor deviations, such as ataxia or stroke. They have not yet been rigorously applied to the fine-grained, subtle MDD characteristic of depression.

Recently, Shen~\etal proposed to rely on Gravity-View Coordinates (GVHMR)~\cite{shen2024gvhmr} to extract 3D human meshes from videos. They propose an optimization approach to enhance the fidelity of the reconstructed meshes; however, the method introduces a significant drift in long videos ($\geq 5$ seconds).

\subsection{The Interpretability Gap in Affective Computing}
Even if we disregard hardware limitations, current computational methods often fail to meet clinical transparency requirements. While state-of-the-art approaches largely rely on end-to-end Convolutional Neural Networks (CNNs)~\cite{suhara2017deepmood, song2018human, icprwdepression2021}, it obscures the specific attributes driving the prediction.

Daoudi \etal~\cite{daoudiaciiw2019} proposed robust motion representations grounded in Riemannian geometry for depression assessment using Kinect depth sensors. While their geometric framework demonstrated clinical relevance, the reliance on specialized hardware limits deployment in standard clinical settings. The Gram matrix trajectory framework of Kacem \etal~\cite{kacem2020gram} provides a scalable geometric alternative. To bridge the gap between geometric rigor and routine psychiatry, we focus on extracting explicit biomechanical parameters—such as ankle dorsiflexion and pelvis tilt—that map directly to PMR semiology, benchmarking against this framework (Section~\ref{sec:results}).

\begin{figure*}[!htbp]
    \centering
    \includegraphics[width=0.95\textwidth]{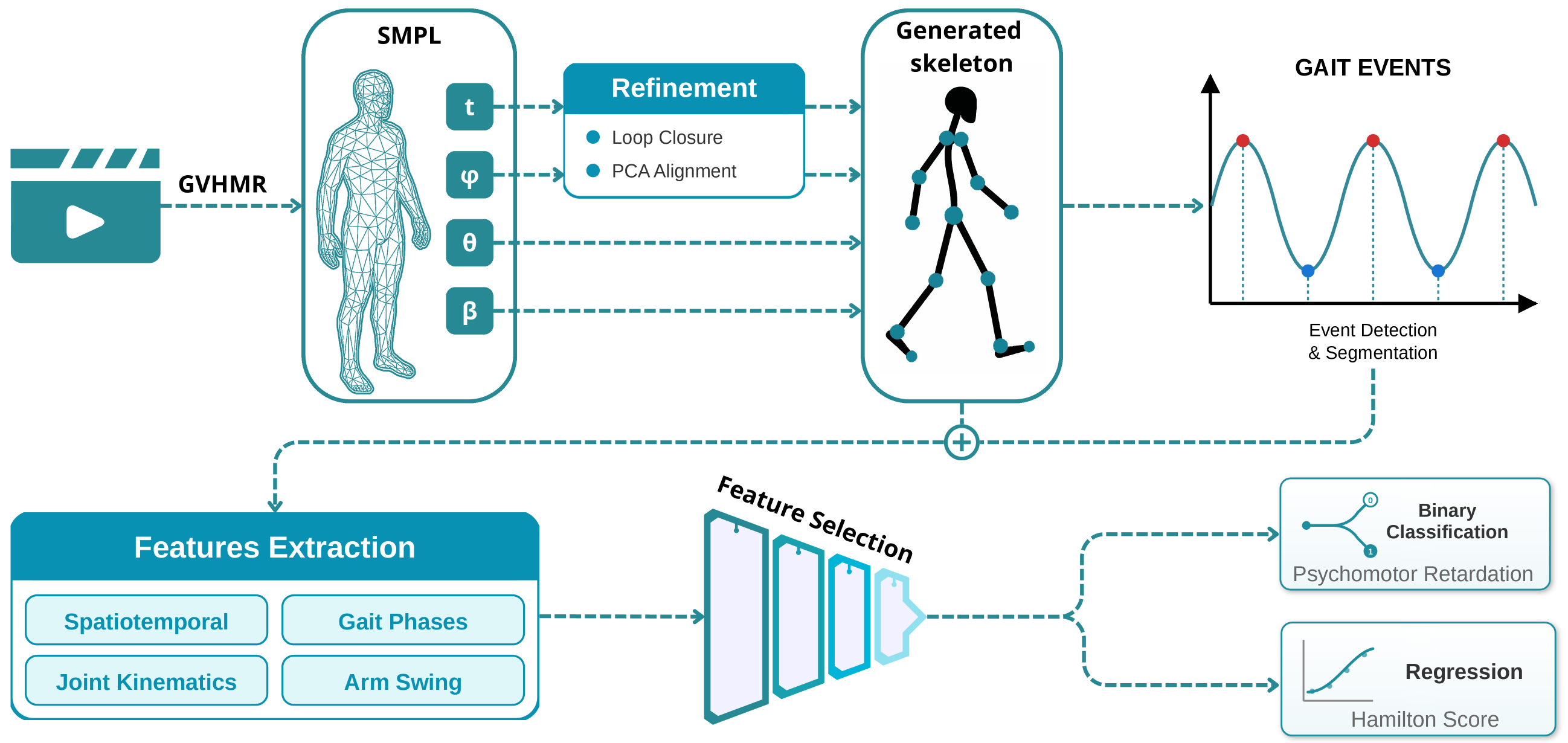}
    \caption{Overview of the proposed pipeline. Monocular video is processed through GVHMR to recover SMPL body parameters: shape ($\beta$), pose ($\theta$), global orientation ($\phi$), and translation ($t$). Trajectory refinement (loop closure, PCA alignment) corrects $\phi$ and $t$ while preserving $\beta$ and $\theta$, producing coherent 3D skeleton sequences. Gait event detection identifies walking phases and arm swing cycles. Extracted features undergo stability-based selection for PMR classification or severity regression.}
    \label{fig:pipeline}
\end{figure*}

\vspace{10pt}
Our method proposes a novel framework for objective analysis of depression and, in particular, for quantification of PMR from monocular RGB video (Fig.~\ref{fig:pipeline}). By relying on a single Timed Up and Go, our protocol is non-invasive, allows for drift correction, and is generalizable to new environments. Moreover, relying on monocular 3D human body estimation, our approach enables the extraction of 297 interpretable biomechanical biomarkers from the protocol, and our stability-based feature selection isolates robust signatures of depression state and PMR on the CALYPSO dataset.

%%%%%%%%%%%%%%%%%%%%%%%%%%%%%%%%%%%%%%%%%%%%%%%%%%%%%%%%%%%%%%%%%%%%%%%%%%%%%%%%

%%%%%%%%%%%%%%%%%%%%%%%%%%%%%%%%%%%%%%%%%%%%%%%%%%%%%%%%%%%%%%%%%%%%%%%%%%%%%%%%
\section{DATASET \& PROTOCOL}
\label{sec:data}

\subsection{Study Cohort}
The study cohort comprises 42 inpatients (22 males, 20 females; mean age $52.7 \pm 15.7$ years) who satisfy the inclusion criteria from the CALYPSO Depression Dataset. All participants were hospitalized for Severe MDD as defined by the DSM-5 criteria~\cite{APA2013DSM5}. Comprehensive demographic and clinical characteristics are summarized in Table~\ref{tab:demographics}.

\begin{table}[!htbp]
\centering
\caption{Demographic and clinical characteristics of the study cohort ($N=42$).}
\label{tab:demographics}
\begin{tabular}{ll}
\hline
\textbf{Characteristic} & \textbf{Value} \\
\hline
\hline
Age (years) & 52.7 $\pm$ 15.7 (range: 20--82) \\
\hline
\textbf{Gender} & \\
\quad Male (M) / Female (F) & 22 (52.4\%) / 20 (47.6\%) \\
\hline
\textbf{HDRS Total Score} & 18.5 $\pm$ 4.7 (range: 6--28) \\
\hline
\textbf{Psychomotor Retardation (Item 8)} & \\
\quad Score 0 (Asymptomatic) & 22 (52.4\%) \\
\quad Score 1--3 (Symptomatic) & 20 (47.6\%) \\
\hline
\textbf{By Group} & \textit{Asympt. / Sympt.} \\
\quad Gender (M/F) & 10/12 \,/\, 12/8 \\
\quad Age (years) & 53.8$\pm$15.7 \,/\, 51.6$\pm$16.1 \\
\quad HDRS Score & 18.2$\pm$5.4 \,/\, 18.9$\pm$4.0 \\
\hline
\end{tabular}
\end{table}

Prior to the motor acquisition, each participant underwent a semi-structured clinical interview conducted by a trained psychiatrist. During this session, the \textit{Hamilton Depression Rating Scale} (HDRS) was administered to establish the clinical ground truth. The clinician scored each item based on a synthesis of verbal responses and behavioral observations. Immediately following the interview, participants performed the gait protocol to ensure that the captured motor signatures corresponded temporally to the current clinical state.\\
The CALYPSO clinical trial has been reviewed and approved
by the Ethics Committee under approval number 2022-
A01160-43, ensuring adherence to ethical standards and
guidelines. All patients provided written informed consent
before participation in the study.
\subsection{Standardized Motor Protocol}
\begin{figure}[!b]
    \centering
    \includegraphics[width=0.90\columnwidth]{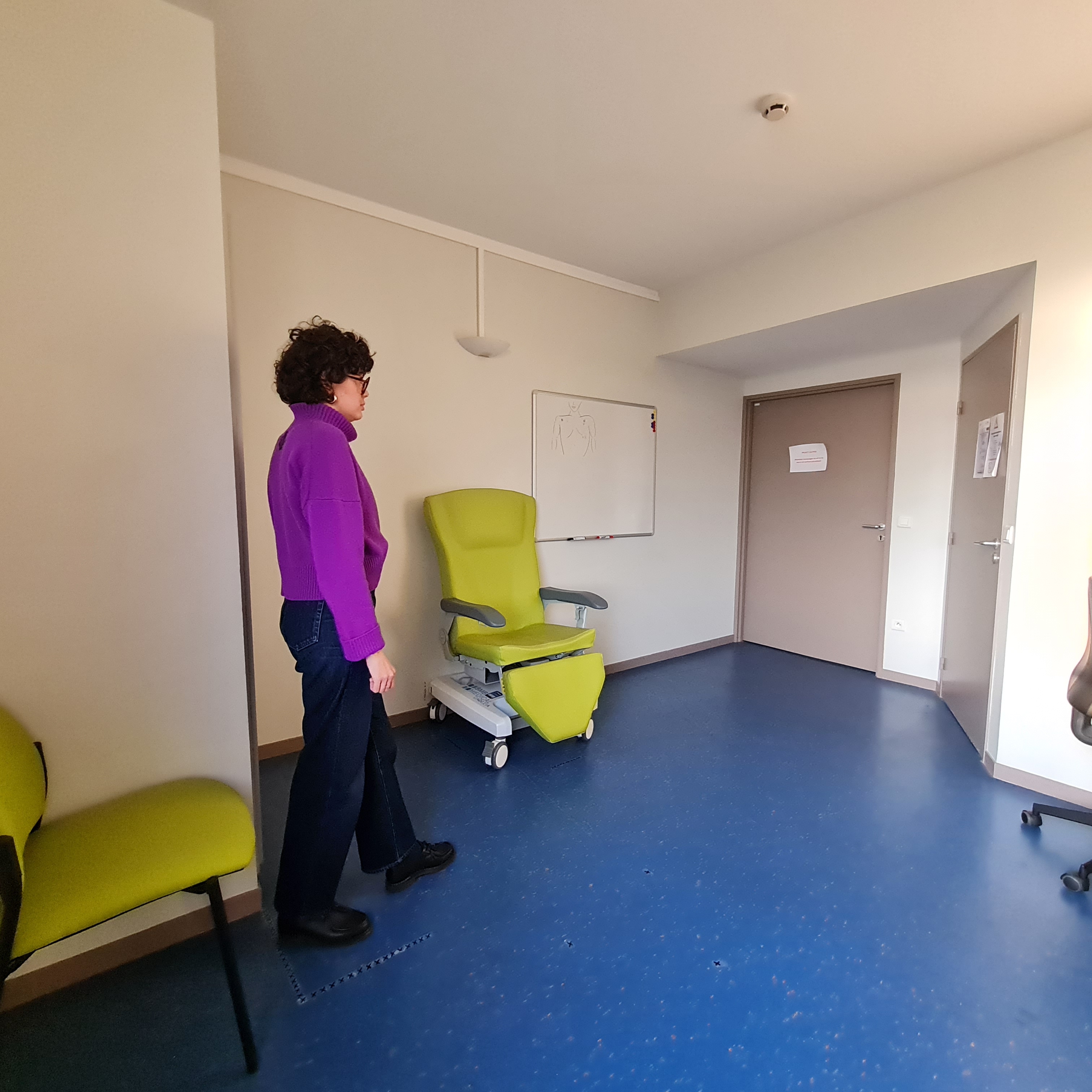}\\[4pt]
    \includegraphics[width=0.95\columnwidth]{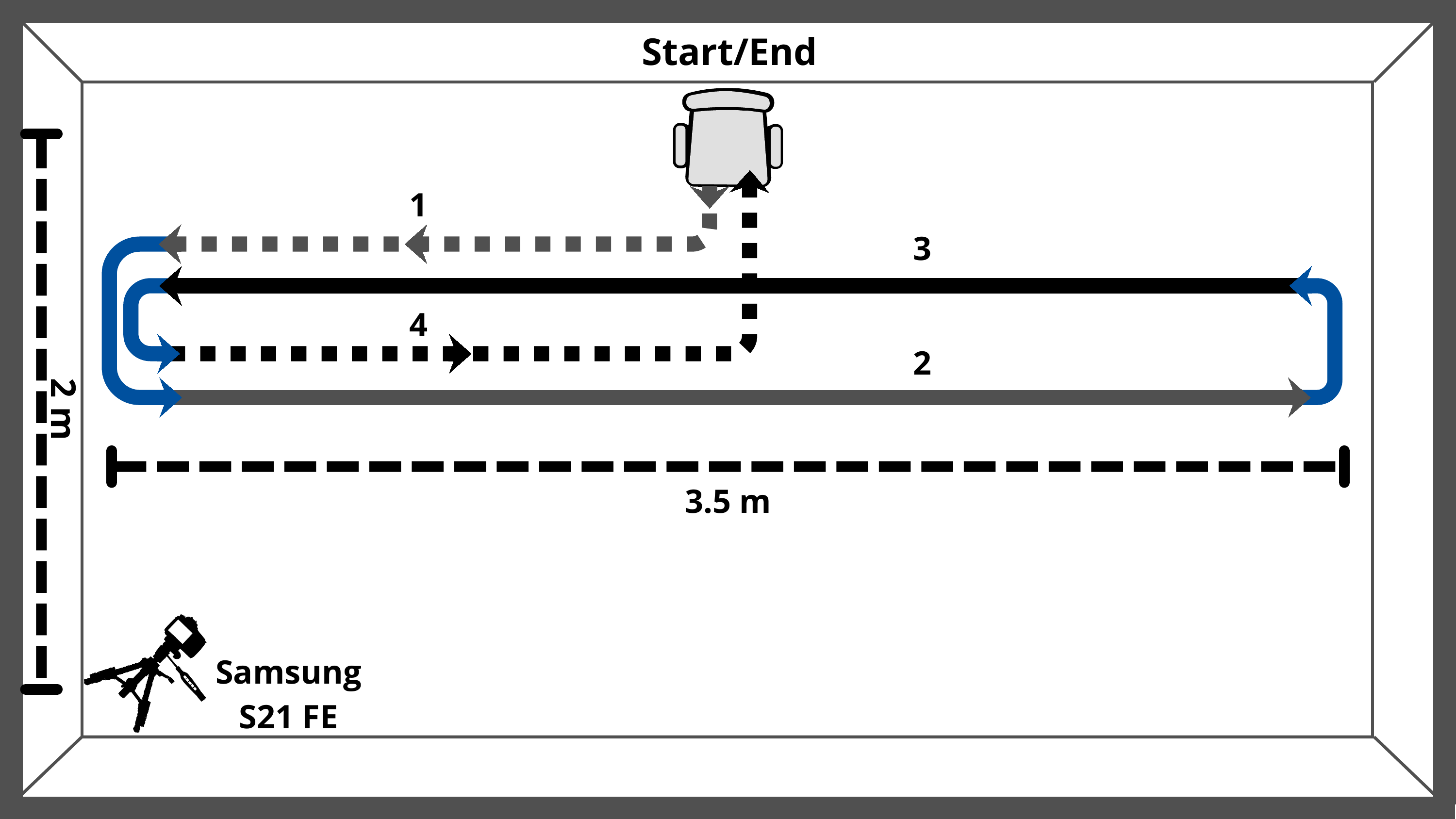}
    \caption{Experimental setup. \textbf{Top:} Camera view of the clinical recording environment. Videos were captured using a Samsung Galaxy S21 FE smartphone (Android 13) at $1440\times1440$ resolution (1:1 aspect ratio) and 30~fps using the ultrawide lens. The device was stabilized on a tripod at a height of approximately 1.5~m. \textbf{Bottom:} Diagram of the modified Timed Up and Go (TUG) protocol, illustrating the four linear movement phases separated by turning segments, designed to ensure that the full TUG trajectory remains within the camera frame.}
    \label{fig:setup_room}

\end{figure}

 To capture standardized gait kinematics, we employed a modified Timed Up and Go (TUG) protocol (Fig.~\ref{fig:setup_room}). Adherence to pathing instructions was strictly monitored; participants exhibiting significant deviations from the protocol were excluded from the final dataset. The motion is captured with a commercial-grade smartphone camera (Samsung Galaxy S21 FE).

The standardized sequence consists of four primary movement phases:% 1) \textbf{Initiation:} The participant stands and moves to the designated start line, 2) \textbf{Outward Walk:} A linear walk of approximately 3.5 meters along a predefined track, 3) \textbf{Return Walk:} A 180$^{\circ}$ turn followed by a linear walk back to the starting point, and 4) \textbf{Termination:} A final turn to face the chair and returning to a seated position.
    
\begin{enumerate}
   \item \textbf{Initiation:} Rising from the chair, the participant walks to the starting position shown in the top view of Fig.~\ref{fig:setup_room}.
    \item \textbf{Outward Walk:} A linear walk of approximately 3.5 meters along a predefined track.
   \item \textbf{Return Walk:} A 180$^{\circ}$ turn followed by a linear walk back to the starting point.
   \item \textbf{Termination:} A final turn to face the chair and returning to a seated position.
\end{enumerate}
This structured protocol explicitly interleaves turning maneuvers between linear displacements, enabling the isolation of steady-state walking segments (Phases 2 and 3) from the complex biomechanics of standing, sitting, and turning. 

\subsection{Prediction Targets}
We utilized the HDRS scores to define two primary prediction objectives:
\begin{itemize}
    \item \textbf{Psychomotor Retardation (Classification):} A binary target derived from HDRS Item 8. Participants were categorized into \textit{asymptomatic} (Score 0, $n=22$) and \textit{symptomatic} (Score $\geq$ 1, $n=20$) groups.
    \item \textbf{Depression Severity (Regression):} The total HDRS score was treated as a continuous metric representing global illness severity.
\end{itemize}

\subsection{Technical Data Pre-processing}

\textbf{1. 3D Mesh Recovery:}
While state-of-the-art methods typically regress SMPL parameters~\cite{loper2015smpl} from monocular video~\cite{kocabas2020vibe}, most operate within root-relative coordinate systems, which are prone to significant global trajectory drift over time. To mitigate this, we adopt the GVHMR framework~\cite{shen2024gvhmr}, which recovers body parameters within a \textbf{Gravity-View coordinate system} aligned with the vertical gravity axis.

For each frame $t$, the framework predicts body shape $\beta \in \mathbb{R}^{10}$, articulated pose $\theta \in \mathbb{R}^{24 \times 3}$, global root orientation $\phi^t \in \mathbb{R}^3$, and root translation $\mathbf{t}^t \in \mathbb{R}^3$. This results in a world-grounded sequence where the floor plane is implicitly defined by the gravity vector, providing a robust foundation for subsequent trajectory optimization and kinematic analysis.

\textbf{2. Trajectory Optimization via Loop Closure:}
Monocular motion capture is susceptible to integration drift, where small frame-to-frame errors accumulate over time~\cite{shen2024gvhmr}. 
To correct this, we exploit the closed-loop TUG topology ($\mathbf{p}^N \approx \mathbf{p}^0$) and formulate a constrained least-squares optimization inspired by SLAM loop-closure techniques~\cite{strasdat2010scale,durrant2006slam}. 
The optimization enforces start-to-end consistency while preserving local motion profiles (velocity preservation)~\cite{gleicher1998retargeting}. 
Details of the algorithm, including an animation, are provided in the supplementary material.

\textbf{3. Coordinate System Standardization:}
While the camera remains stationary, the raw reconstructed trajectory reflects the participant's specific heading, which seldom aligns perfectly with the camera's optical axis. To ensure consistency across the cohort, we rigidly align the TUG walkway with the global $Z$-axis.

Given that the $Y$-axis is already aligned with gravity via GVHMR, we determine the principal direction of progression by computing the first eigenvector $\mathbf{v}_1 = (v_x, v_z)$ of the trajectory covariance matrix projected onto the horizontal $XZ$-plane. The yaw rotation $\theta_{\text{align}}$ required to align this vector with the global $Z$-axis is computed as:
\begin{equation}
    \theta_{\text{align}} = -\operatorname{atan2}(v_x, v_z).
\end{equation}
The rotation matrix $\mathbf{R}_y(\theta_{\text{align}})$ is then applied to the global translation $\mathbf{t}^t$ and the root orientation $\phi^t$ for the entire sequence:
\begin{equation}
    \mathbf{t}^{t}_{\text{aligned}} = \mathbf{R}_y(\theta_{\text{align}})\,\mathbf{t}^t, \quad \phi^{t}_{\text{aligned}} = \mathbf{R}_y(\theta_{\text{align}}) \circ \phi^t.
\end{equation}

This transformation effectively decouples longitudinal displacement from lateral sway, ensuring that extracted kinematic features remain invariant to the participant's specific pathing angle relative to the camera lens.

\section{PROPOSED FRAMEWORK}
Our preprocessing pipeline described above generates a 3D skeletal sequence $\mathbf{S} = \{\mathbf{J}^t\}_{t=1}^{N}$. A single $\mathbf{J}^t \in \mathbb{R}^{45 \times 3}$ contains the positions of 45 joints at frame $t$, comprising the standard SMPL skeletal joints augmented with high-fidelity facial and foot landmarks. 

This rectified skeletal representation, corrected for integration drift and aligned to the principal walking axis, provides a consistent spatiotemporal basis for the automated extraction of clinical biomarkers.

\subsection{Temporal Segmentation}
The TUG protocol consists of linear walking phases interspersed with 180$^{\circ}$ turns (see~Fig.~\ref{fig:setup_room}). To isolate stable gait cycles, the recording is first segmented by identifying these turning maneuvers. Since the trajectory is already aligned with the principal walking axis (see Section~\ref{sec:data}), the segmentation task is reduced to a 1D signal analysis. Specifically, we define the progression signal $s(t)$ as the pelvis coordinate projected along the standardized forward axis ($Z$).

\begin{figure}[tb]
    \centering
    \includegraphics[width=\columnwidth]{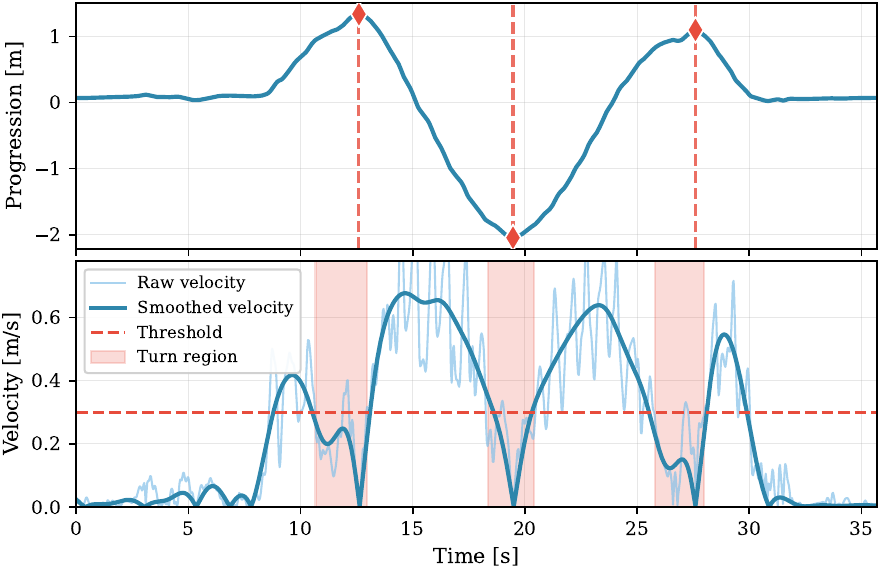}
    \caption{Turn detection during the TUG test. Top: Progression along the principal walking axis with detected turn centers (red diamonds). Bottom: Velocity profile showing raw (light blue) and smoothed (dark blue) signals. Turns are identified when velocity falls below the threshold (dashed red line), with turn regions shaded.}
    \label{fig:turn_detection}
\end{figure}

\textit{Turn Detection.}
Turns are identified as directional transitions within this 1D progression signal. As illustrated in Fig.~\ref{fig:turn_detection}, we employ a two-stage detection strategy: initial candidate identification via peak detection on $s(t)$, followed by precise boundary refinement based on velocity dynamics.

\textit{Candidate Identification:} 
The centers of the turning maneuvers correspond to the reversal points of the out-and-back trajectory. These are detected as local extrema (maxima and minima) within the progression signal $s(t)$:
\begin{equation}
    \mathcal{T}_{\text{cand}} = \{\, t \mid s(t) \text{ is a local maximum or minimum}\,\}.
\end{equation}

\textit{Segment Refinement via Velocity Dynamics:} 
To accurately delineate the transitions between walking and turning, we analyze the instantaneous velocity magnitude along the principal axis:
\begin{equation}
    v(t) = \left| \frac{d}{dt} s(t) \right|.
\end{equation}
To suppress high-frequency jitter while preserving locomotor transitions, the velocity signal is smoothed using a second-order Butterworth low-pass filter ($f_c = 1.5$\,Hz) applied with zero-phase filtering. We define a dynamic threshold $\tau$ to identify turning intervals:
\begin{equation}
    \tau = \min\left(0.4\,\text{m/s}, \; 0.5 \times v_{95\%}\right),
\end{equation}
where $v_{95\%}$ represents the 95th percentile of the velocity distribution. Turn phases are identified as contiguous intervals surrounding each candidate point where $v(t) < \tau$. To ensure robustness against momentary hesitations or spurious velocity drops, a minimum duration constraint of 500\,ms is enforced; intervals failing this criterion are discarded. The remaining segments between validated turns are classified as ``Steady-State Walking'' and isolated for kinematic feature extraction.

These thresholds, along with the minimum turn duration of 500\,ms, were determined empirically through visual inspection and fixed for all subsequent analyses.
\subsection{Gait Event Detection}

Within the identified steady-state walking segments, we detect discrete gait events to delineate individual gait cycles. We utilize a kinematic approach based on the relative position of foot landmarks (heel and toe) with respect to the pelvis, adapted from Zeni \textit{et al.}~\cite{zeni2008two}.

Let $z_{\text{heel}}(t)$ and $z_{\text{toe}}(t)$ denote the global $Z$-coordinates (forward axis) of the heel and toe, respectively, and $z_{\text{pelvis}}(t)$ represent the pelvis position. We compute the relative longitudinal displacements as follows:
\begin{align}
    z_{\text{heel,rel}}(t) &= z_{\text{heel}}(t) - z_{\text{pelvis}}(t), \\
    z_{\text{toe,rel}}(t) &= z_{\text{toe}}(t) - z_{\text{pelvis}}(t).
\end{align}

\begin{figure}[!htbp]
    \centering
    \includegraphics[width=\columnwidth]{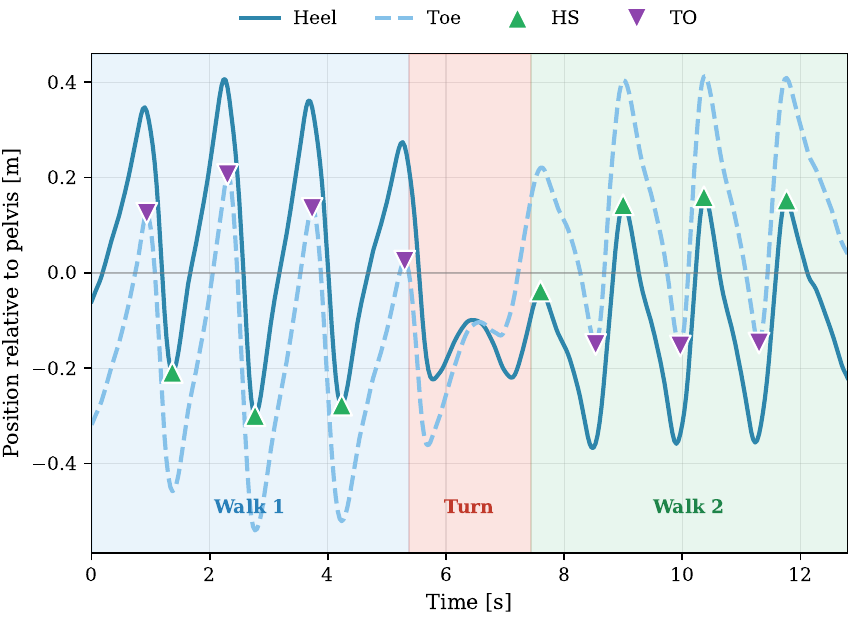}
    \caption{Gait event detection during the TUG protocol. Left foot heel (solid) and 
    toe (dashed) trajectories relative to the pelvis are shown. Green upward triangles 
    indicate Heel Strikes (HS); purple downward triangles denote Toe-Offs (TO). 
    Background colors distinguish between protocol phases: blue (outward walking), 
    red (turning), and green (returning).}
    \label{fig:gait_events}
\end{figure}

The event detection logic is illustrated in Fig.~\ref{fig:gait_events}. The walking direction $\hat{d}_z \in \{+1, -1\}$ is determined by the net pelvis displacement within each segment. For segments where $\hat{d}_z = +1$, Heel Strikes (HS) are identified as local maxima in $z_{\text{heel,rel}}(t)$ (signifying maximal forward extension), while Toe-Offs (TO) correspond to local minima in $z_{\text{toe,rel}}(t)$ (signifying maximal backward extension). For return walking segments ($\hat{d}_z = -1$), these extrema criteria are inverted.

A single \textbf{gait cycle} is defined as the interval between two consecutive ipsilateral heel strikes ($\text{HS}_i$ to $\text{HS}_{i+1}$). This interval is subdivided into a \textit{stance phase} ($\text{HS}_i \to \text{TO}_i$) and a \textit{swing phase} ($\text{TO}_i \to \text{HS}_{i+1}$), which serve as the temporal windows for subsequent kinematic feature extraction.

\subsection{Arm Swing Dynamics}
To isolate arm dynamics from global locomotion, we define a pelvis-attached local coordinate system using Gram-Schmidt orthonormalization~\cite{winter2009biomechanics}. The \textbf{up axis} $\hat{\mathbf{u}}$ is defined by the vector from the pelvis to the upper spine (spine3), the \textbf{side axis} $\hat{\mathbf{s}}$ is the vector from the right hip to the left hip, and the \textbf{forward axis} is computed via the cross product $\hat{\mathbf{f}} = \hat{\mathbf{s}} \times \hat{\mathbf{u}}$, ensuring a mutually orthogonal basis. The wrist position relative to the pelvis is then projected onto this local frame:
\begin{align}
    d_{\text{fwd}}(t) &= (\mathbf{p}_{\text{wrist}}(t) - \mathbf{p}_{\text{pelvis}}(t)) \cdot \hat{\mathbf{f}}, \\
    d_{\text{lat}}(t) &= (\mathbf{p}_{\text{wrist}}(t) - \mathbf{p}_{\text{pelvis}}(t)) \cdot \hat{\mathbf{s}},
\end{align}
where $d_{\text{fwd}}$ and $d_{\text{lat}}$ capture the longitudinal (forward-backward) and lateral swing components, respectively.

While GVHMR recovers full-body meshes regardless of direct visibility, predictions for occluded segments may exhibit reduced reliability. To ensure data integrity, we flag frames where wrist detection confidence falls below 0.85 and exclude the arm from analysis if occlusions exceed 50\% of the recording duration. From the remaining high-confidence signals, swing cycles are segmented by identifying local extrema in $d_{\text{fwd}}(t)$; maxima represent peak forward extension, while minima denote peak backward extension. A complete swing cycle is defined as the interval between consecutive minima.

\subsection{Joint Angle Computation}

Joint angles are computed directly from recovered 3D positions following International Society of Biomechanics (ISB) conventions~\cite{wu2002isb}, avoiding artifacts from SMPL rotation parameters. We compute angles for:
\begin{itemize}
    \item \textbf{Lower limb:} knee flexion/extension, hip flexion and abduction, ankle dorsiflexion/plantarflexion.
    \item \textbf{Pelvis:} sagittal tilt, frontal obliquity, transverse rotation.
    \item \textbf{Trunk:} forward lean, lateral lean, trunk-pelvis dissociation.
\end{itemize}

\begin{figure}[!t]
    \centering
    \includegraphics[width=0.9\columnwidth]{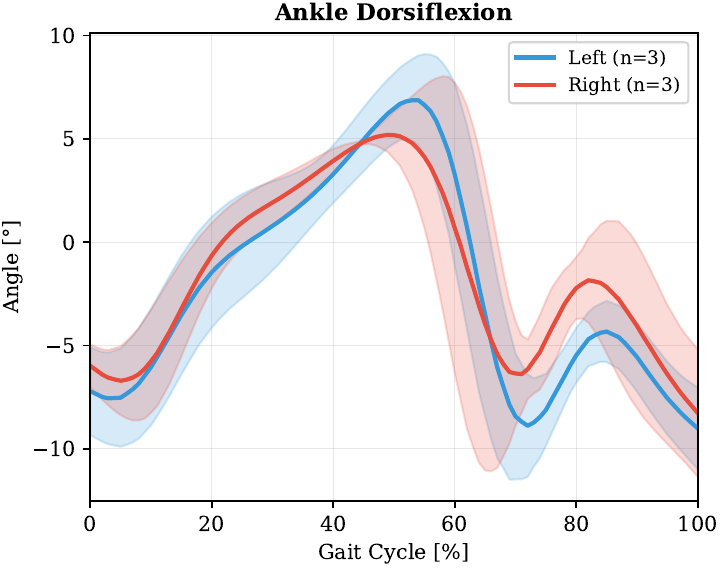}
    \caption{Ankle dorsiflexion across normalized gait cycles for a representative participant. Shaded regions indicate standard deviation across $n=3$ cycles.}
    \label{fig:ankle_dorsiflexion}
\end{figure}

To enable inter-subject comparison, continuous angle trajectories are time-normalized: each gait cycle is linearly interpolated to a standardized domain spanning $0\%$--$100\%$ of the cycle duration, as illustrated in Fig.~\ref{fig:ankle_dorsiflexion}.

\subsection{Digital Biomarkers}
We extract parameters across six biomechanical domains: spatiotemporal metrics (velocity, cadence, step/stride length), temporal phases (stance, swing, double-support percentages), joint kinematics (value at HS/TO, peak values during stance/swing, timing of peaks, and ROM), arm swing dynamics (amplitude, symmetry), locomotor variability (CV of metrics), and TUG performance (turning duration, angular velocity). Per-cycle metrics are aggregated (mean, SD, CV) and averaged bilaterally, yielding \textbf{297 features}, detailed in the Supplementary Material.

\section{Machine Learning Pipeline}
\label{sec:ml_pipeline}

To derive robust and interpretable gait signatures from our limited cohort ($N=42$), we implemented a two-stage machine learning framework designed to mitigate overfitting while ensuring clinical transparency.

\subsection{Stage 1: Stability-Based Feature Selection}

\noindent Given the high-dimensional nature of our data ($p=297, N=42$), we mitigate overfitting risks using stability selection inspired by~\cite{Meinshausen2010,boualeb2025measuring} (Algorithm~\ref{alg:pipeline}). By selecting features that consistently appear across various subsamples, this method ensures the extraction of stable gait signals while filtering out sampling noise.

\begin{algorithm}[!b]
\caption{Stability-Based Feature Selection}
\label{alg:pipeline}
\begin{algorithmic}[1]
\Require Gait features $\mathbf{X}$, clinical targets $\mathbf{y}$
\Ensure Consensus feature set $\mathcal{S}$

\For{each fold $k = 1, \ldots, K$}
    \State Partition data into training and validation sets
    \State \textbf{Redundancy Filtering:} If $|\rho_{ij}| > 0.9$, retain feature $j$ with higher $|\text{corr}(j, \mathbf{y})|$
    \State \textbf{Wrapper Selection:} Apply SBS with inner CV to minimize validation error.
    \State Store selected features $\mathcal{F}_k$
\EndFor
%\Statex
\State \textbf{Consensus Aggregation:}
\For{each feature $j$}
    \State $\pi_j \gets$ selection frequency across all $K$ folds
\EndFor
\State $\mathcal{S} \gets \{j \mid \pi_j > 0.5\}$, restricted to the top 10 features (EPV constraint)

\State \Return $\mathcal{S}$
\end{algorithmic}
\end{algorithm}

We utilized a 10-fold cross-validation (CV) scheme. Within each fold, the training partition ($N \approx 38$) underwent two sequential processing steps:
\begin{enumerate}
    \item \textbf{Target-Aware Collinearity Filtering:} Features exceeding a pairwise Pearson correlation threshold ($|\rho| > 0.9$) were identified. From each correlated pair, we retained only the feature with the highest absolute correlation to the target variable. This prevents collinear variables from diluting the feature importance signal and ensures a more parsimonious model.
    
    \item \textbf{Sequential Backward Selection (SBS):} A wrapper-based backward elimination was applied to minimize the cross-validated error. The algorithm iteratively removed the least informative features until the performance reached an optimal plateau.
\end{enumerate}

This procedure generates 10 independent feature subsets. We compute a \textit{Stability Score} for each feature, defined as its selection frequency across all iterations. The final \textbf{Consensus Feature Set} retains the parameters with a Stability Score $> 50\%$, capped at a maximum of 10 features to satisfy Events-Per-Variable (EPV) constraints and prevent overfitting given our sample size. We further validated the model through permutation testing (N = 1000 random label shuffles) to generate an empirical null distribution, confirming that the captured clinical signal significantly exceeds chance levels despite the non-nested selection.

%Stability-Based Feature Selection was conducted on the full cohort using stability selection \cite{Meinshausen2010}, as the iterative sub-sampling required by the algorithm made nested LOOCV computationally intractable. To mitigate potential data leakage, we prioritized features with high selection frequency across sub-samples, ensuring the identified markers are robust to individual variations. 

% \subsection{Stage 2: Model Benchmarking and Validation}
% \label{stage2}

% \textbf{Evaluation Protocol:} To maximize training data utilization while providing unbiased performance estimates, we employed \textbf{Leave-One-Out Cross-Validation (LOOCV)}.

% \textbf{Statistical Validation:} To confirm that the observed performance significantly exceeded chance, we performed permutation testing ($N_{\text{perm}} = 1000$). By randomly shuffling patient labels to establish null distributions for accuracy and $R^2$, we assessed the statistical significance of our model's predictive power.

%%%%%%%% i am here i ll continue tomorow%%%%%%%%%%%%%%
\section{Experiments and Results}
\label{sec:results}

\subsection{Experimental details and baselines}

We validated the proposed framework on two distinct clinical tasks: the classification of PMR and the estimation of overall depression severity.  Performance metrics for both tasks are summarized in Table~\ref{tab:performance_summary}. 

Using the fixed Consensus Feature Set, we conducted a systematic benchmark of several architectures, including regularized linear models, ensemble methods, and Support Vector Machines (SVMs). The best performing architectures are Logistic Regression (for classification) and Linear SVR (for regression). All preprocessing steps, including robust scaling, were embedded within the CV loop to strictly prevent data leakage.

For comparison with the literature, we also benchmark against the Gram matrix method~\cite{kacem2020gram}, which has shown solid results for interpretable depression state assessment. We adapt this baseline approach to our extracted skeletons using ppfSVM for classification and SVR for regression (see Supplementary Material).

\subsection{Prediction Performance}

\begin{table}[H]
\centering
\caption{Performance comparison with baseline. Permutation testing ($N=1000$) confirms significance.}
\label{tab:performance_summary}
\begin{tabular}{|l|l|c|c|c|}
\hline
\textbf{Task} & \textbf{Method} & \textbf{\# Feat.} & \textbf{Performance} & \textbf{$p$-value} \\
\hline
\hline
Regression & Gram Matrix~\cite{kacem2020gram} & -- & $R^2$: 0.13 & 0.03 \\
 & \textbf{Proposed} & 9 & \boldmath{$R^2$: 0.64} & $<$0.001 \\
\hline
Classification & Gram Matrix~\cite{kacem2020gram} & -- & Acc: 0.67 & 0.10 \\
 & \textbf{Proposed} & 7 & \textbf{Acc: 0.83} & $<$0.001 \\
\hline
\end{tabular}
\end{table}

The framework's performance for severity estimation and PMR detection is summarized in Table~\ref{tab:performance_summary}.

\textbf{Regression (Depression Severity):}
The model explained 64\% of the variance ($R^2=0.64$) using 9 features: arm swing dynamics (forward duration (s), frequency (Hz), and lateral-to-forward ratio variability), lower limb kinematics (peak ankle dorsiflexion during stance ($^\circ$), peak knee extension timing (\% cycle) and amplitude ($^\circ$)), and pelvis mobility (tilt amplitude ($^\circ$), rotation timing (\% cycle)).

\textbf{Classification (Psychomotor Retardation):} 
The system detected PMR with 83.3\% accuracy with only 7 features spanning ankle kinematics (dorsiflexion at heel strike ($^\circ$), toe-off variability (\%), peak variability (\%), and minimum ($^\circ$)), arm lateral swing amplitude variability (\%), peak pelvic rotation variability (\%), and minimum hip abduction variability (\%).

\textbf{Benchmark Comparison and Statistical Validation:}
Our explicit markers achieved high significance across both tasks (Table~\ref{tab:performance_summary}) compared to the Gram matrix trajectory framework \cite{kacem2020gram}, which struggled to capture clinical severity and failed to reach statistical significance for PMR detection. This contrast, confirmed by permutation testing, demonstrates that for small clinical cohorts, explicit biomechanical markers provide a more robust and predictive signal than high-dimensional, implicit geometric representations, which appear more susceptible to overfitting.

\subsection{Ablation Study: Impact of Trajectory Refinement}

To quantify the contribution of each algorithmic component, we conducted a systematic ablation study. All configurations utilized identical feature extraction and machine learning pipelines, varying only the trajectory preprocessing steps (Table~\ref{tab:ablation}).

\begin{table}[t]
    \centering
    \caption{Ablation study evaluating the effect of trajectory refinement components.}
    \label{tab:ablation}
    \resizebox{\linewidth}{!}{%
    \begin{tabular}{|l|c|c|cc|}
    \hline
     & \multicolumn{2}{c|}{\textbf{Classification}} & \multicolumn{2}{c|}{\textbf{Regression}} \\
    \cline{2-5}
    \textbf{Configuration} & \textbf{Acc.} & \textbf{AUC} & \boldmath{$R^2$} & \textbf{RMSE} \\
    \hline
    \hline
    \textbf{Full Pipeline} & \textbf{0.83} & \textbf{0.89} & \textbf{0.64} & \textbf{2.78} \\
    \hline
    w/o Loop Closure & 0.76 & 0.85 & 0.63 & 2.81 \\
    \hline
    w/o Velocity Smoothing & 0.76 & 0.78 & 0.56 & 3.08 \\
    \hline
    w/o PCA Alignment & 0.73 & 0.76 & 0.47 & 3.39 \\
    \hline
    GVHMR Only (Baseline) & 0.73 & 0.75 & 0.22 & 4.10 \\
    \hline
    \end{tabular}%
    }
\end{table}

The results demonstrate that the full refinement pipeline is essential for clinical validity. For classification, \textit{loop closure} was the most critical factor (reducing accuracy to 76.2\% when removed). For regression, \textit{PCA alignment} proved decisive; without it, $R^2$ dropped from 0.64 to 0.47. Notably, using raw GVHMR without any refinement degraded regression performance by 66\%, confirming that geometric constraints are required to transform monocular output into clinically meaningful biomarkers.

%To provide a geometric baseline for both classification and regression, we adapted the Gram matrix trajectory framework of Kacem \etal~\cite{kacem2020gram} to operate on our GVHMR-extracted skeletons, using ppfSVM for classification and SVR for regression (see Supplementary Material for implementation details). As shown in Table~\ref{tab:performance_summary}, the proposed framework significantly outperformed this baseline across both tasks.

%Linear SVR substantially outperformed the Gram Matrix baseline (Table~\ref{tab:performance_summary}). This improvement demonstrates the clinical value of explicit biomechanical features over implicit geometric representations. The model utilized 9 features capturing three domains: arm swing dynamics (forward duration (s), frequency (Hz), and lateral-to-forward ratio variability), lower limb kinematics (peak ankle dorsiflexion during stance ($^\circ$), peak knee extension timing (\% cycle) and amplitude ($^\circ$)), and pelvis mobility (tilt amplitude ($^\circ$), rotation timing (\% cycle)).

%The framework identified PMR with high accuracy, surpassing the Gram Matrix baseline (Table~\ref{tab:performance_summary}). The selected 7 features spanned ankle kinematics (dorsiflexion at heel strike ($^\circ$), toe-off variability (\%), peak variability (\%), and minimum ($^\circ$)), arm lateral swing amplitude variability (\%), peak pelvic rotation variability (\%), and minimum hip abduction variability (\%).

\subsection{Biomechanical Interpretability}
\begin{figure}[!t]
    \centering
    \setlength{\tabcolsep}{1pt}
    \begin{tabular}{c}
         \includegraphics[width=0.8\columnwidth]{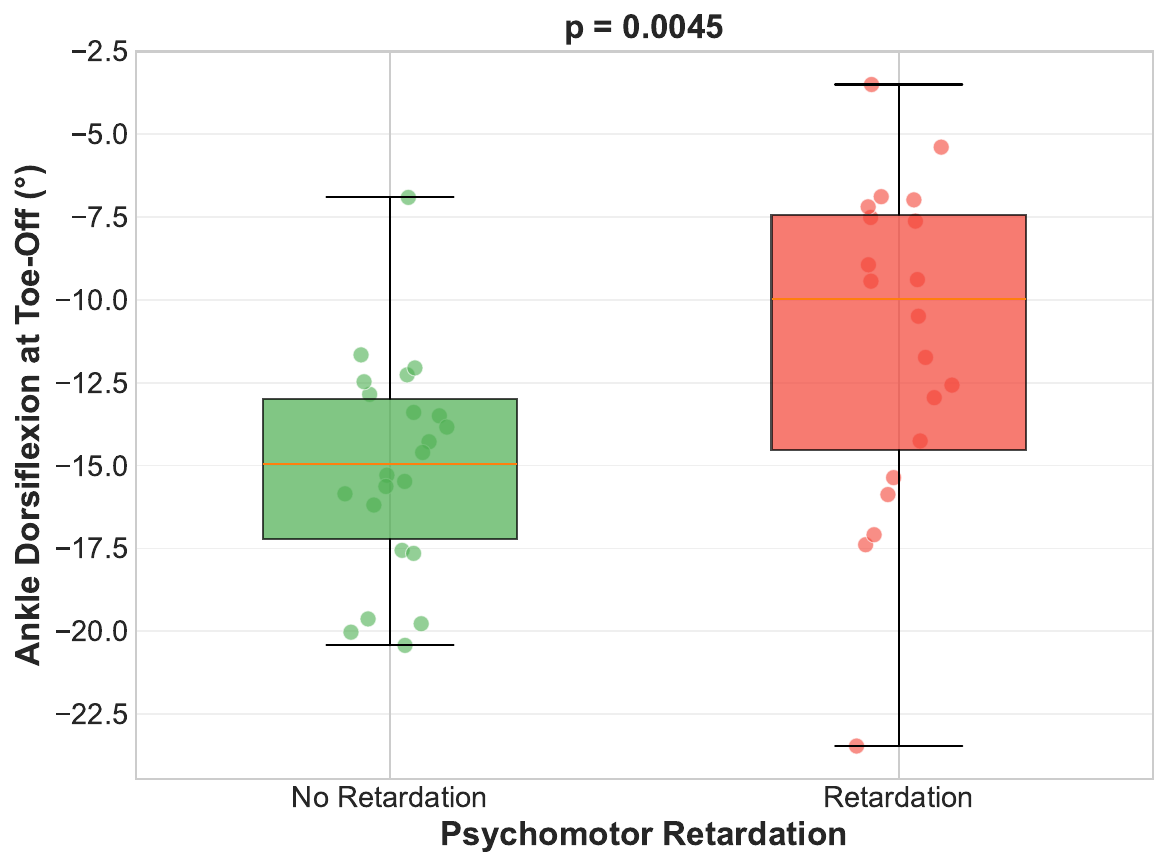} \\
         {\footnotesize (a) Ankle Dorsiflexion at Toe-Off by Retardation Status} \\[4pt] 
         \includegraphics[width=0.8\columnwidth]{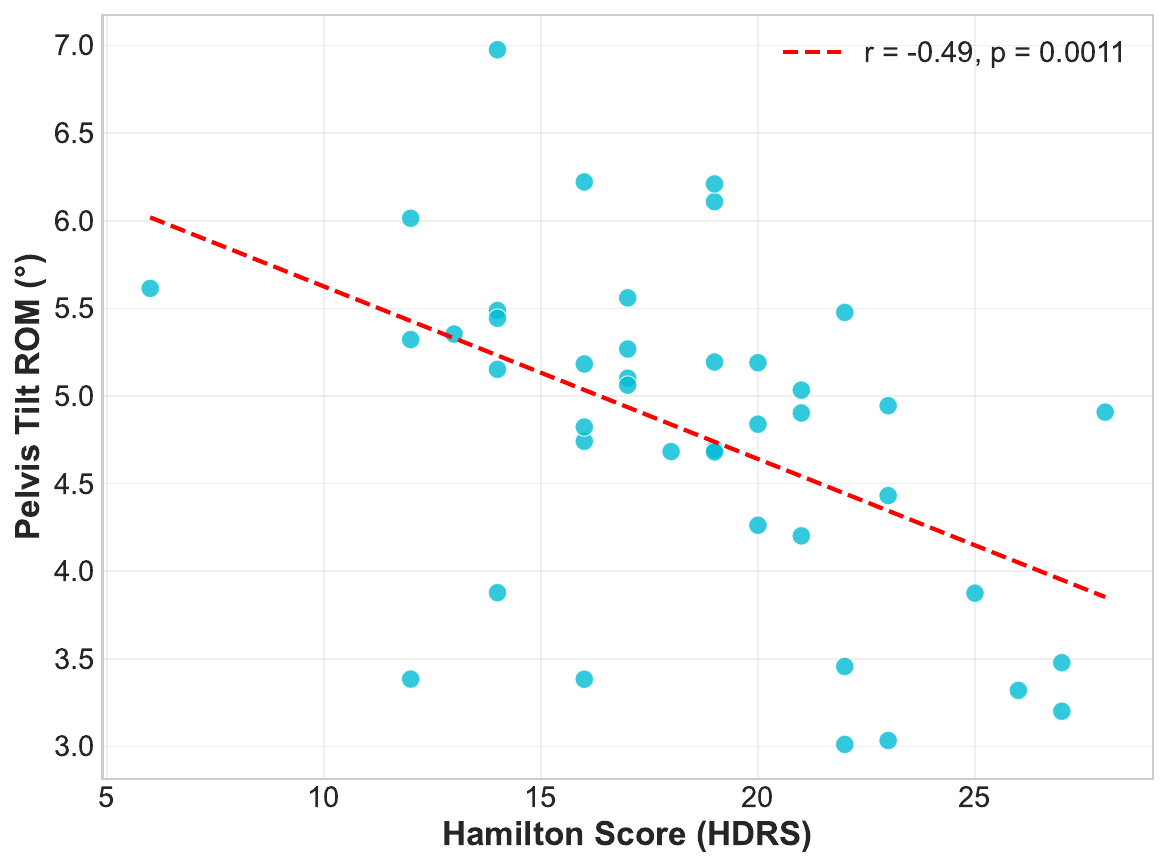} \\
         {\footnotesize (b) Pelvis Tilt ROM vs. Depression Severity}
    \end{tabular}
    \caption{Univariate analysis. (a) PMR patients show reduced ankle plantarflexion (weaker push-off). (b) Restricted pelvic mobility correlates with depression severity (axial rigidity).}
    \label{fig:feature_correlations}
\end{figure}

Stability selection prioritized features from \textbf{ankle propulsion} and \textbf{axial dynamics}. While the model leveraged a combination of spatiotemporal and variability metrics for performance, we analyze clinical signatures using mean values and Range of Motion (ROM) in Fig.~\ref{fig:feature_correlations} to facilitate interpretability:

\begin{itemize}
    \item \textbf{Ankle Propulsion:} Fig.~\ref{fig:feature_correlations}(a) shows that patients with PMR exhibit significantly reduced mean plantarflexion. This captures the less energetic push-off phase identified by the model, serving as a measurable kinetic deficit of psychomotor output.
    
    \item \textbf{Axial Rigidity:} Beyond orientation variability, Fig.~\ref{fig:feature_correlations}(b) shows pelvic tilt ROM correlates negatively with depression severity. This restricted mobility objectively captures the ``stiffening'' of the body axis typical of the depressive phenotype.
\end{itemize}

%%%%%%%%%%%%%%%%%%%%%%%%%%%%%%%%%%%%%%%%%%%%%%%%%%%%%%%%%%%%%%%%%%%%%%%%%%%%%%%%
\section{Discussion}

\subsection{Gait as a Proxy for Cognitive State}
A significant finding is the 83.3\% accuracy achieved for HDRS Item 8. This item represents a clinician's global assessment of psychomotor slowing, primarily capturing perceived cognitive latencies—such as slowness of thought and speech—during the clinical interview. The fact that our model predicted this rating solely from gait kinematics suggests that physical locomotion serves as a robust, observable proxy for this cognitive state.

Our framework effectively leverages specific kinetic anomalies—notably reduced ankle push-off and axial rigidity—as an objective readout of psychomotor processing speed. These findings substantiate the functional coupling between motor and cognitive slowing in depression, pointing to a shared neurobiological dysfunction within the striato-frontal circuits which simultaneously govern locomotor control and executive function. By measuring the physical breakdown in these circuits through gait, we effectively capture the parallel breakdown in central executive processing.

\subsection{The Upper Bound of Motor-Based Severity Estimation}
While PMR is directly measurable, total HDRS aggregates 17 symptoms (e.g., guilt, insomnia), which sometimes do not show motor manifestations. However, gait alone explains 64\% of variance, confirming that psychomotor dysfunction is central to the disorder, not peripheral. Our study delineates a natural ``upper bound'' on gait analysis: it effectively quantifies the somatic burden but cannot fully capture emotional dimensions.

\section{Conclusion}
This study presents a novel, non-invasive framework for the objective quantification of PMR in patients with MDD. Our findings demonstrate that physical movement constitutes a robust indicator of depressive pathology. By automatically extracting 297 biomarkers, we successfully detected PMR with 83.3\% accuracy and accounted for 64\% of the variance in overall depression severity. Specific motor signatures, such as reduced ankle push-off and pelvic instability, were identified as key indicators of MDD pathology. These findings suggest that movement analysis can serve as a robust, scalable digital biomarker for depression severity. By providing interpretable signatures, this approach bridges the gap between non-invasive motion-capture facilities and routine psychiatric practice.

By relying on gait data acquired through a simple and standardized task, objective motion-based assessments may offer scalable and clinically applicable measures of psychomotor functioning, compatible with routine care and longitudinal monitoring.

\section{Limitations and Future Work}

Despite these promising results, several limitations warrant consideration. First, the modest sample size (N=42) may constrain the generalizability of our findings; however, the robustness of our effects was substantiated by rigorous permutation testing. Second, gait kinematics is frequently influenced by confounders—including age, BMI, and pharmacological factors such as drug-induced parkinsonism. Future iterations should mitigate these influences through larger, stratified cohorts or the integration of adversarial debiasing techniques.

Furthermore, we utilized a single metric (HDRS Item 8) to operationalize PMR. While clinically meaningful, it may not capture the construct's full multidimensionality. Subsequent research should evaluate this framework against dedicated instruments, such as the CORE measure~\cite{parker1994defining} or Salpêtrière Retardation Rating Scale (SRRS)~\cite{widlocher1983psychomotor}, to determine if motion features can serve as a sensitive proxy for these specific assessments.

From a technical perspective, markerless mesh recovery remains sensitive to loose-fitting clothing, which is common in clinical populations. Additionally, the computational complexity of the GVHMR framework currently precludes real-time feedback; future research could employ knowledge distillation to enable deployment on edge devices. Finally, the standardized TUG protocol may induce a Hawthorne effect, where patients alter their gait under observation. To improve ecological validity, future studies should explore unobtrusive monitoring in naturalistic environments. Additionally, integrating gait with facial and speech biomarkers through multimodal fusion could offer a more holistic and robust digital phenotype for depression.

%%%%%%%%%%%%%%%%%%%%%%%%%%%%%%%%%%%%%%%%%%%%%%%%%%%%%%%%%%%%%%%%%%%%%%%%%%%%%%%%
\section*{ACKNOWLEDGMENTS}
The French State under the France-2030 programme and the Initiative of Excellence of the University of Lille are acknowledged for the funding and support granted to the R-CDP-24-005-CALYPSO project.
%%%%%%%%%%%%%%%%%%%%%%%%%%%%%%%%%%%%%%%%%%%%%%%%%%%%%%%%%%%%%%%%%%%%%%%%%%%%%%%%
% 
% This year, at FG 2026, we will be introducing a new requirement that authors submit an Ethical Impact Statement as part of the submission process. (Note that this requirement applies to all short and long papers submitted to the main track. Each special track may have its own rules.) To support authors in meeting this requirement, as well as reviewers in assessing it, we have prepared this document with general guidelines, a checklist for authors and reviewers to complete, and answers to some frequently asked questions. As this is a new policy, we welcome questions and feedback as we refine it and develop a shared understanding.
%\clearpage
%\section*{ETHICAL IMPACT STATEMENT}
%\input{FG2026_Latex_template/ethicalstatement}
%\clearpage
\section*{Supplementary Materials}
In this supplementary document, we provide details about the recording setup and the trajectory optimization algorithm, and we also include a table summarizing the biomechanics parameters. 
%% ============================================
%\section{Recording Setup}

%Videos were recorded using a Samsung Galaxy S21 FE smartphone (Android 13) at 1440$\times$1440 resolution (1:1 aspect ratio) and 30\,fps. The ultrawide lens (0.5$\times$, $\sim$120° FOV) was used to capture the full TUG trajectory within the frame. The device was mounted on a tripod at approximately 1.5\,m height.

%% ============================================
\section{Trajectory Optimization}
\subsection{Optimization Algorithm}

We use a constrained least-squares optimization inspired by SLAM loop-closure~\cite{strasdat2010scale,durrant2006slam} to enforce TUG start-to-end consistency while preserving local motion profiles~\cite{gleicher1998retargeting}. 

Let $\mathbf{p}_t$ denote the pelvis position at frame $t$. We minimize:
\begin{equation}
\mathbf{p}^* = \arg\min_{\mathbf{p}^*} \; J_{\text{smooth}} + J_{\text{loop}}
\quad \text{subject to} \quad \mathbf{p}^*_0 = \mathbf{p}_0
\end{equation}
where the two energy terms enforce complementary objectives:

\begin{itemize}
    \item \textbf{Velocity Preservation:} Maintains the local motion profile by penalizing deviations from the observed frame-to-frame displacements $\Delta \mathbf{p}_t = \mathbf{p}_{t+1} - \mathbf{p}_t$:
    \begin{equation}
    J_{\text{smooth}} = \sum_{t=0}^{T-2}
    \left\| (\mathbf{p}^*_{t+1} - \mathbf{p}^*_t) - \Delta \mathbf{p}_t \right\|^2
    \end{equation}

    \item \textbf{Loop Closure:} Corrects accumulated drift by enforcing that the final position matches the starting point:
    \begin{equation}
    J_{\text{loop}} = \left\| \mathbf{p}^*_{T-1} - \mathbf{p}^*_0 \right\|^2
    \end{equation}
\end{itemize}

Both terms are weighted equally in the optimization. The hard constraint $\mathbf{p}^*_0 = \mathbf{p}_0$ anchors the trajectory at the known starting location, preventing the optimizer from uniformly shifting the entire path.

We solve this sparse linear system efficiently by decoupling spatial dimensions. While patients may have slight deviations returning to the chair ($\approx$10--20\,cm), these are negligible relative to monocular integration drift ($>$0.5\,m), making closed-loop enforcement a valid approximation.

%\subsection{Animations}

%An animation is provided showing two synchronized skeletons: the original trajectory (with accumulated drift) and the corrected trajectory after loop closure optimization. This side-by-side comparison illustrates how the optimization preserves the walking pattern while eliminating positional drift.

\section{Feature Extraction Methodology}
\subsection{Projection onto Walking Direction}

Spatiotemporal metrics are computed relative to the \textit{instantaneous walking direction} to ensure accuracy during turns. The walking direction $\hat{\mathbf{d}}$ is defined as the unit vector of pelvis displacement between consecutive heel strikes:
\begin{itemize}
    \item \textbf{Step Length:} $|\,(\mathbf{h}_{\text{land}} - \mathbf{h}_{\text{trail}}) \cdot \hat{\mathbf{d}}\,|$ (projection onto $\hat{\mathbf{d}}$)
    \item \textbf{Step Width:} $|\,(\mathbf{h}_{\text{land}} - \mathbf{h}_{\text{trail}}) \cdot \hat{\mathbf{d}}_\perp\,|$ (projection onto perpendicular)
\end{itemize}
This ensures that step length captures only the forward component and step width captures only the lateral component, even during turning phases.

\subsection{Extracted Parameters}
Table~\ref{tab:parameters} provides a comprehensive overview of the 297 biomechanical parameters, categorized into six domains: spatiotemporal metrics, temporal phases, lower limb kinematics, pelvis and trunk kinematics, arm swing dynamics, and functional variability measures.

\begin{table*}[!t]
\centering
\caption{Comprehensive Summary of Gait, Kinematic, and Arm Swing Parameters}
\label{tab:parameters}
\footnotesize
\begin{tabularx}{\textwidth}{llX}
\toprule
\textbf{Category} & \textbf{Parameter} & \textbf{Definition / Convention / Units} \\ \midrule
\textbf{Spatiotemporal} & Step-Level & Length (m), Width (m), Time (s), Velocity (m/s), Height (m), Pelvis Progression (m) \\
& Stride-Level & Stride Length (m), Stride Time (s), Cadence (strides/min) \\
& Normalized & Step Length/Leg Length; Velocity/$\sqrt{g \cdot \text{Leg Length}}$ (Froude) \\ \midrule
\textbf{Temporal Phases} & Stance \& Swing & Duration (s) and \% of Gait Cycle (GC) \\
& Double Support & \% of GC where both feet are in contact with ground \\ \midrule
\textbf{Lower Limb} & Knee Flexion & Extension = 0, Flexion = (+) [$^\circ$] \\
(Kinematics) & Hip Flexion & Thigh forward = (+) [$^\circ$] \\
& Hip Abduction & Leg away from midline = (+) [$^\circ$] \\
& Ankle DF & Dorsiflexion = (+), Plantarflexion = (--) [$^\circ$] \\ \midrule
\textbf{Pelvis \& Trunk} & Pelvis & Tilt (Ant+), Obliquity (L-Up+), Rotation (L-Fwd+) [$^\circ$] \\
(Kinematics) & Trunk & Flexion (Lean+), Lateral Lean (L+), Rotation (Dissociation) [$^\circ$] \\
& Metrics & Values at HS/TO, Peaks, Timing (\%), and ROM \\ \midrule
\textbf{Arm Swing} & Amplitude & Forward, Backward, and Lateral wrist displacement (m) \\
(Pelvis-frame) & Temporal & Half-cycle duration (s), Period (s), Frequency (Hz) \\
& Classification & Normal ($\geq$5cm), Rigid ($<$5cm), Parasite (Lat/Fwd ratio $>$0.5) \\ \midrule
\textbf{Functional \&} & TUG Turns & Turn Duration (s), Peak Turn Angular Velocity (deg/s) \\
\textbf{Variability} & Variability & CV (\%) of Step L/T/V, Stride T, and Step Width \\ \bottomrule
\end{tabularx}
\end{table*}

\section{Baseline Implementation Details}

To benchmark our explicit biomechanical markers against a state-of-the-art geometric approach, we adapted the Gram matrix trajectory framework of Kacem \etal~\cite{kacem2020gram}. While the original framework supports both 2D and 3D data, we adapted it specifically for 3D skeletal landmarks ($d=3$) to match the output of our GVHMR system.

\textbf{Geometric Representation:} 
Each frame $t$ of the skeletal sequence is represented as a Gram matrix $G_t = Z_t Z_t^\top$, where $Z_t \in \mathbb{R}^{n \times 3}$ is the mean-centered landmark matrix. 
The matrix $G_t$ is decomposed via polar decomposition $Z_t = U_t R_t$, yielding a subspace component $U_t \in \mathcal{G}(3, n)$ on the Grassmannian and a strictly positive-definite energy matrix $P_t = R_t^2 \in \mathcal{P}_3$ on the SPD manifold.

\textbf{Temporal Alignment \& Metric Learning:} 
Pairwise trajectory distances are computed using Dynamic Time Warping (DTW) with the \textit{split metric}:
\begin{equation}
    d_{\mathcal{S}^+}^2(G_1, G_2) = d_{\mathcal{G}}^2(U_1, U_2) + k \cdot d_{\mathcal{P}_3}^2(P_1, P_2),
\end{equation}
where $d_{\mathcal{G}}$ is the Grassmannian geodesic distance (via principal angles) and $d_{\mathcal{P}_3}$ is the Affine-Invariant Riemannian metric on SPD matrices. 
We used hyperparameters reported as optimal for 3D skeletal data: $k = 0.25$ and temporal resampling thresholds $\zeta_1, \zeta_2$ set to normalize trajectory length~\cite{kacem2020gram}.

\textbf{Learning Tasks:} 
Following the Pairwise Proximity Function SVM (ppfSVM) formulation~\cite{kacem2020gram}, we constructed a feature vector $\phi(\beta_G)$ for each trajectory containing its pairwise DTW distances to all training samples. These vectors were used to train a linear SVM, which effectively linearizes the Riemannian manifold structure for classification.

\begin{itemize}
    \item \textbf{Classification (Psychomotor Retardation):} Linear SVM trained on proximity vectors, with hyperparameters optimized via cross-validation. Validation was performed using Leave-One-Out Cross-Validation (LOSO-CV).
    
    \item \textbf{Regression (Depression Severity):} Linear SVR trained on proximity vectors. Performance was evaluated via LOSO-CV and statistical significance was verified using permutation testing ($N=1000$).
\end{itemize}

\section{Detailed Model Comparison}

To identify optimal consensus features, we applied stability-based selection~\cite{Meinshausen2010} using multiple embedded models (Lasso, Ridge, ElasticNet, SVR) across stability thresholds from 20\% to 100\%. For regression, Lasso ($\alpha=0.1$) at 50\% threshold achieved the highest $R^2=0.65$, yielding 9 consensus features. For classification, Logistic Regression (L1, C=1) at 50\% threshold achieved the highest accuracy (83.3\%), yielding 7 consensus features. Lower thresholds included noisy features, while higher thresholds (>80\%) often produced empty feature sets.

Table~\ref{tab:model_comparison} presents performance across different model families using Leave-One-Out Cross-Validation with proper scaling inside each fold to prevent data leakage. Linear models consistently outperformed non-linear approaches (Random Forest, KNN, SVC-RBF), suggesting that the relationship between gait features and clinical outcomes is predominantly linear.

\begin{table}[t]
\centering
\caption{Model Comparison on Consensus Feature Sets}
\label{tab:model_comparison}
\footnotesize
\begin{tabular}{lcc}
\toprule
\multicolumn{3}{c}{\textbf{Classification (7 features)}} \\
\textbf{Model} & \textbf{Acc.} & \textbf{AUC} \\
\midrule
\textbf{Logistic Reg (L1, C=1)} & \textbf{0.83} & 0.89 \\
Ridge ($\alpha$=1) & 0.81 & \textbf{0.91} \\
Logistic Reg (L2, C=1) & 0.79 & 0.86 \\
SVC (Linear, C=1) & 0.76 & 0.82 \\
SVC (Linear, C=0.1) & 0.74 & 0.75 \\
Random Forest & 0.67 & 0.73 \\
KNN (k=5) & 0.64 & 0.69 \\
\midrule
\multicolumn{3}{c}{\textbf{Regression (9 features)}} \\
\textbf{Model} & \textbf{$R^2$} & \textbf{RMSE} \\
\midrule
\textbf{SVR (Linear, C=2)} & \textbf{0.64} & \textbf{2.78} \\
SVR (Linear, C=1) & 0.64 & 2.80 \\
Ridge ($\alpha$=1) & 0.54 & 3.15 \\
ElasticNet ($\alpha$=0.1) & 0.54 & 3.15 \\
Lasso ($\alpha$=0.1) & 0.54 & 3.16 \\
Random Forest & 0.34 & 3.79 \\
KNN (k=5) & 0.27 & 3.98 \\
\bottomrule
\end{tabular}
\end{table}

%%%%%%%%%%%%%%%%%%%%%%%%%%%%%%%%%%%%%%%%%%%%%%%%%%%%%%%%%%%%%%%%%%%%%%%%%%%%%%%%
%{\small
%\bibliographystyle{ieee}
%\bibliography{egbib}
%}
%%%

\end{document}